\documentclass[11pt]{article}

\usepackage[margin=1in]{geometry}
\usepackage{amsmath,amssymb,amsthm}
\usepackage{natbib}
\usepackage{graphicx}
\usepackage{url}
\usepackage{algorithm2e}
\usepackage{longtable}
\usepackage{booktabs}
\usepackage{siunitx}
\usepackage{multirow}

\title{Reason Popper-ly: Patching In-Context Reasoning with Inductive Logic Programming}
\author{%
  Zirong Chen \qquad Meiyi Ma\\[0.5em]
  \normalsize Department of Computer Science, Vanderbilt University, USA\\
  \normalsize\texttt{zirong.chen@vanderbilt.edu \qquad meiyi.ma@vanderbilt.edu}%
}
\date{\textit{Accepted at the 20th Conference on Neurosymbolic Learning and Reasoning}}

\begin{document}

\maketitle

\begin{abstract}
Chain-of-thought (CoT) prompting enables large language models (LLMs) to tackle multi-step reasoning tasks, yet the generated intermediate steps are not guaranteed to be logically sound. We present Reason \textsc{Popper}-ly, a neurosymbolic framework that uses inductive logic programming (ILP) to learn relation composition rules from reasoning traces and deploys them as an online verifier for step-level correction. Given an LLM-generated trace, the method checks each inferred step against the learned rule table, diagnoses the violation type, rewrites incorrect steps with symbolically derived repairs, and regenerates the remaining suffix so that the model can produce its final answer conditioned on a verified trace. We evaluate on CLUTRR, a multi-hop kinship reasoning benchmark, using five language models over reasoning chains of 2 to 10 hops. Across all models, Reason \textsc{Popper}-ly consistently improves terminal accuracy over standard CoT, with gains of up to 48 percentage points for small models and 15 points for frontier models on the longest chains. Compared with a fully exogenous symbolic pipeline, our method performs better on harder instances by preserving the model's successful grounding while correcting only verifiable reasoning failures. In addition, step-level ILP verification yields a fine-grained error taxonomy that provides diagnostic insight beyond final-answer accuracy.
\end{abstract}

\section{Introduction}

Chain-of-thought (CoT) prompting has become a standard approach for eliciting multi-step reasoning from large language models (LLMs) \cite{wei2022chain}. By generating intermediate reasoning steps before producing a final answer, LLMs often achieve strong gains on tasks that require composition, deduction, and planning. However, stronger end-task performance does not imply that the generated reasoning trace is logically sound or faithful to the model's actual decision process. Prior work shows that CoT explanations can be plausible while remaining unfaithful \cite{turpin2023language,lanhammeasuring}, and that reasoning quality degrades substantially as compositional depth increases \cite{saparov2023language}. These findings point to a gap between the surface form of structured reasoning and its underlying validity.
A natural response is to combine LLMs with symbolic methods. Recent neurosymbolic approaches either translate natural-language problems into formal representations and delegate reasoning to external solvers \cite{pan2023logic,olausson2023linc}, or inject symbolic structure into the reasoning process to improve faithfulness and controllability \cite{xu2024faithful,lyu2023faithful}. These approaches show that symbolic constraints can improve reasoning reliability, but they typically operate at the level of the full problem or the full reasoning trace. In contrast, we study a more targeted setting: using symbolic knowledge as an online guardrail over the model's own reasoning trace, preserving valid intermediate work while intervening at steps that violate explicit compositional rules.

Bridging this gap with symbolic methods raises several open \textbf{\textit{challenges}}:
(1) \textit{Rule acquisition}. Symbolic verification requires explicit inference rules. Classical inductive logic programming (ILP) \cite{muggleton1991inductive,cropper2021learning} can learn such rules from examples rather than requiring manual specification, but it has rarely been used as a runtime verifier for LLM-generated reasoning traces. Existing neurosymbolic approaches either internalize rules through retraining with logic-informed objectives \cite{de2025inductive}, or rely on hand-crafted formal grammars specified per domain \cite{pan2023logic,chen2022cityspec,chen2023cityspec}. Neither directly learns verification rules from model-generated reasoning traces and deploys them for online step-level intervention.
(2) \textit{Grounding bottleneck}. Symbolic verification operates over structured predicates, whereas CoT traces are written in natural language. This creates a grounding bottleneck in which extraction or translation errors can compound with reasoning errors. Logic-LM \cite{pan2023logic} partially addresses this issue by using LLMs as translators from natural language to formal logic, but the translation stage itself introduces an additional failure mode. More broadly, this bottleneck has historically limited symbolic reasoning pipelines to relatively narrow and well-formalized domains \cite{olausson2023linc,lyu2023faithful}.
(3) \textit{Error attribution}. Current neurosymbolic verifiers often reduce verification to a binary accept-or-reject decision over the full trace or full solution. SymbCoT \cite{xu2024faithful} integrates symbolic expressions into chain-of-thought with an LLM-based verification stage, but does not localize failures to specific reasoning steps or systematically classify their types. A practically useful verifier should identify \textit{which} step failed and \textit{why}, such as a wrong composition rule, a missing inversion, a direction reversal, or a hallucinated premise. Such step-level attribution is important for targeted intervention and aligns with the broader motivation for process-level reasoning supervision \cite{lightman2023let}.
(4) \textit{Targeted correction}. Retry-based refinement methods \cite{madaan2023self,gou2024critic} often discard the entire context and regenerate from scratch, losing correct intermediate work. Replacing a single step is more targeted, but it also changes the context for all subsequent steps. The corrected trace must therefore remain coherent, and the model must condition on the patched prefix in a way that meaningfully improves the final answer. This is especially important because intrinsic self-correction without reliable external feedback remains limited for reasoning tasks \cite{huang2024large}.

Targeting these challenges, we propose Reason \textsc{Popper}-ly, a framework that learns relation-composition rules from reasoning traces via ILP and applies them as an online verifier over CoT reasoning. Given a complete reasoning trace, our method checks each inferred step against an induced composition table, classifies detected violations, replaces invalid steps with symbolically grounded corrections, and regenerates only the remaining suffix conditioned on the patched prefix. This design preserves the model's own reasoning whenever it is already sound, while introducing symbolic correction only where it is needed.

Our \textbf{\textit{contributions}} are:
(1) We propose Reason \textsc{Popper}-ly, an ILP-based pipeline for step-level verification and in-place repair of LLM reasoning traces. Rather than replacing model reasoning with a standalone symbolic solver, our method uses learned symbolic rules as an online guardrail over the model's own CoT.
(2) We introduce an error taxonomy for compositional reasoning failures that provides fine-grained diagnostic information beyond binary correctness. The taxonomy is enabled by the structured nature of ILP-based verification and supports localized attribution of reasoning failures.
(3) We evaluate our method on CLUTRR \cite{sinha2019clutrr} across multiple language models and reasoning depths. The results show that step-level symbolic patching consistently improves terminal accuracy over standard CoT, and becomes increasingly advantageous over a fully exogenous symbolic pipeline as reasoning depth grows.
\section{Background and Preliminaries}

We briefly review in-context reasoning and inductive logic programming that are used throughout the paper.

\subsection{In-Context Reasoning}
 
Given a pretrained language model $\mathcal{M}$ and an input sequence $x$ consisting of a task description and optional demonstrations, \textit{in-context reasoning} refers to the process by which $\mathcal{M}$ generates a response $y$ through conditional next-token prediction, without updating model parameters. The model relies  on the information available within its context window.
 
\paragraph{Chain-of-thought (CoT).}
Chain-of-thought prompting \cite{wei2022chain} elicits intermediate reasoning steps $s_1, s_2, \ldots, s_n$ before a final answer $a$. We denote the resulting reasoning trace by
\[
\tau = (s_1, \ldots, s_n, a).
\]
Each step $s_i$ is a natural-language statement that either (i) restates information explicitly given in the input, which we call a \textit{stated} step, or (ii) derives new information from previous steps, which we call an \textit{inferred} step.

\paragraph{Step soundness.}
Let $\mathcal{R}$ be a set of inference rules. An inferred step $s_i$ is \textit{sound} with respect to $\mathcal{R}$ if the conclusion expressed by $s_i$ follows from its cited premises under some rule $r \in \mathcal{R}$. A reasoning trace $\tau$ is sound if every inferred step in $\tau$ is sound. Soundness is necessary but not sufficient for final-answer correctness, since a trace may still begin from incorrectly extracted or misinterpreted premises.

\subsection{Inductive Logic Programming}
 
Inductive Logic Programming (ILP) \cite{muggleton1991inductive} learns a logic program $H$ from background knowledge $B$, positive examples $E^+$, and negative examples $E^-$ such that:
\begin{align}
    B \cup H &\models E^+ \label{eq:completeness} \\
    B \cup H &\not\models E^- \label{eq:consistency}
\end{align}
That is, the learned hypothesis $H$ together with $B$ should entail all positive examples while remaining consistent with all negative examples.
In our setting, the learned hypothesis consists of ground facts over a ternary predicate $\texttt{compose}/3$. Formally, $\texttt{compose}(r_1, r_2, r_3),$
where $r_1, r_2, r_3 \in \mathcal{V}$ are relation symbols drawn from a finite vocabulary $\mathcal{V}$. Each fact states that composing relation $r_1$ with relation $r_2$ yields relation $r_3$. Collectively, these facts define a partial composition table $\mathcal{R}: \mathcal{V} \times \mathcal{V} \rightharpoonup \mathcal{V}.$

We use \textsc{Popper} \cite{cropper2021learning}, an ILP system based on learning from failures\footnote{Codebase: \url{https://github.com/logic-and-learning-lab/Popper}}. Popper follows a generate, test, and constrain loop: it proposes candidate hypotheses using answer set programming, evaluates them against the examples in Prolog, and derives constraints from failed candidates to prune subsequent search. In our case, this allows us to induce a compact symbolic rule table that can later be used for runtime verification of LLM reasoning steps.

\section{Method: Reason \textsc{Popper}-ly}

\begin{figure}[t]
    \centering
    \includegraphics[width=0.6\linewidth]{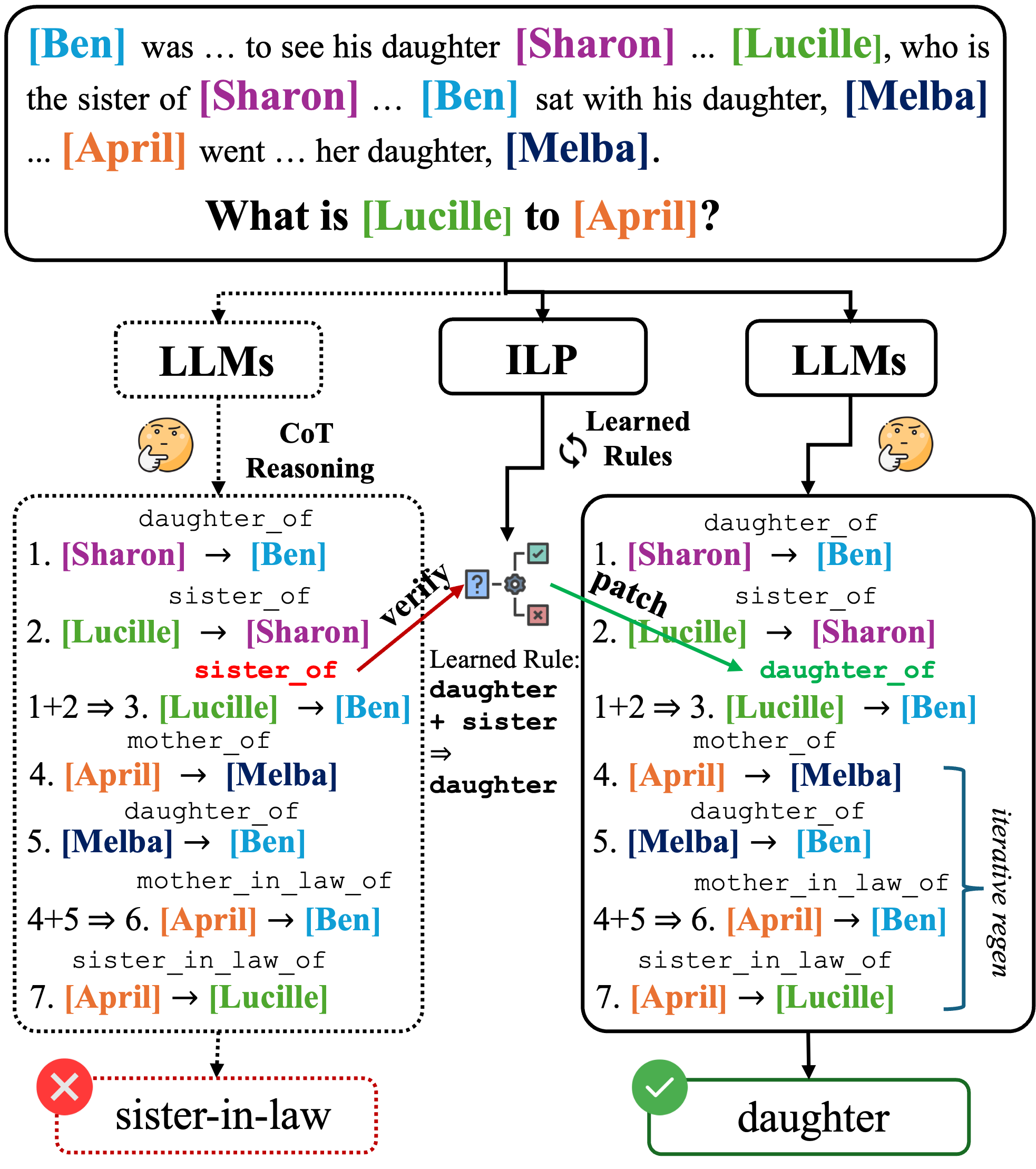}
    \caption{Overview of Reason \textsc{Popper}-ly. Left: an LLM-generated reasoning trace may contain locally plausible but compositionally invalid steps. Right: our method detects violated steps, diagnoses the error type, patches the offending step, and regenerates the remaining suffix conditioned on the corrected prefix.}
    \label{fig:example}
\end{figure}

We propose Reason \textsc{Popper}-ly, a framework that uses ILP-learned relation-composition rules to verify and repair LLM chain-of-thought traces at the step level. The framework has two phases. In the offline phase, we induce a symbolic composition table from reasoning traces. In the online phase, we apply this table as a runtime verifier that detects, diagnoses, and repairs invalid inferred steps in newly generated traces. Figure~\ref{fig:example} gives an overview.

\subsection{Offline: Learning Composition Rules}
\label{sec:offline}
 
This offline phase constructs a frozen composition table $\mathcal{R}$ that maps valid relation pairs to their composed relation. This table is learned once and reused for all test instances.

\paragraph{Trace collection.}
We prompt each LLM with structured CoT instructions on a training set of CLUTRR stories. For each instance, the model produces a reasoning trace
$
\tau = (s_1, \ldots, s_n, a),
$
where each step explicitly indicates whether it is \textit{stated} or \textit{inferred}, together with the indices of the supporting prior steps. We then partition the collected traces by terminal correctness. Traces whose final answer matches the gold label contribute positive supervision, while traces with incorrect final answers contribute negative supervision.

\paragraph{Composition extraction.}
For every inferred step that cites exactly two prior steps, we extract a composition triple $(r_1, r_2, r_3)$, which encodes the claim that relation $r_1$ composed with relation $r_2$ yields relation $r_3$. This extraction requires \textit{direction normalization}: relations expressed in the story may not align with the direction needed by the reasoning chain. We therefore map each relation into chain-consistent form using a gender-aware inverse map when necessary. For example, if the story states ``Mildred is the daughter of Gerald,'' the corresponding relation from Gerald to Mildred in the reasoning chain is \texttt{father}.

\paragraph{Rule induction.}
We provide the extracted positive and negative triples to \textsc{Popper} \cite{cropper2021learning} as examples of the target predicate $\texttt{compose}/3$. Particularly designed for kinship reasoning, the hypothesis space is restricted to ground facts of the form $\texttt{compose}(r_1, r_2, r_3)$, so that each learned clause corresponds to a single entry in the composition table rather than a higher-order rule schema. The induced facts are then validated against a gold kinship composition table derived from CLUTRR \cite{sinha2019clutrr} and supplemented with any missing gold entries. The resulting table $\mathcal{R}$ serves as the symbolic verifier used at inference time.

\subsection{Online: Verify, Diagnose, Patch}
\label{sec:online}
 
At inference time, the LLM first produces a complete reasoning trace. We then apply a post-hoc verification-and-repair procedure to this trace. We do not intervene during autoregressive decoding itself, which keeps the procedure simple and avoids complications related to hidden-state or key-value cache manipulation. 

\paragraph{Step 1: Verify.}
Given a trace $\tau = (s_1, \ldots, s_n, a)$, we parse each inferred step into a composition triple $(r_1, r_2, r_3)$ using the same extraction and normalization procedure as in the offline phase. We then check the triple against the learned composition table $\mathcal{R}$. If $(r_1, r_2) \in \mathrm{dom}(\mathcal{R})$ but $\mathcal{R}(r_1, r_2) \neq r_3$, the step is marked as a violation. If the relation pair is outside the domain of $\mathcal{R}$, the step is treated as unverifiable and left unchanged.

\paragraph{Step 2: Diagnose.}
Once violation is detected, we assign it one of four diagnostic types:
(1) \texttt{wrong\_rule}: the model uses the correct pair of premise relations but predicts the wrong composed relation;
(2) \texttt{missing\_inv}: the model fails to invert a premise relation into the chain-consistent direction before composition;
(3) \texttt{direction\_rev}: the model predicts the inverse of the correct target relation;
(4) \texttt{hallucinated}: the model cites a premise relation that is not supported by the story or prior stated facts.
This classification is obtained by comparing model-produced triple against rule-table entry together with the inverse-relation map. The resulting taxonomy provides more information than binary verification alone: it not only identifies that a step is invalid, but also exposes the structural reason for failure.

\paragraph{Step 3: Patch.}
For each step $s_i$, we define a verified step $s_i^*$ according to whether the step is constraint-consistent. Let $\mathcal{C}$ denote the set of constraint-consistent steps. Then:
\begin{equation}
s_i^* =
\begin{cases}
s_i, & \text{if } s_i \in \mathcal{C} \\
\textsc{Patch}(s_i; \mathcal{R}), & \text{otherwise}
\end{cases}
\end{equation}
where $\textsc{Patch}(\cdot)$ is a constraint-aware repair operator that rewrites the relation in $s_i$ using the corrected composition $\mathcal{R}(r_1, r_2)$ while preserving the involved entities and step context as much as possible.
If $s_i \notin \mathcal{C}$ and later steps are entailed from such $s_i$, we replace $s_i$ with $s_i^*$ and discard all subsequent steps. The remaining suffix is then regenerated conditioned on the patched prefix and diagnosed errors, yielding
$
\tau' = (s_1, \ldots, s_{i-1}, s_i^*, \ldots, s_n').
$
We apply this repair process iteratively until no further verifiable violations remain, resulting in a fully verified trace
$
\tau^* = (s_1^*, \ldots, s_n^*).
$
Finally, the repaired trace $\tau^*$ is returned to the LLM as context for generating the final answer.
Conceptually, this procedure acts as a local repair operator over the reasoning trace: it preserves the model's original reasoning wherever possible, intervenes only at rule-violating steps, and restores downstream coherence by regenerating only the affected suffix. 

\section{Evaluation}

We evaluate on CLUTRR \cite{sinha2019clutrr}, a benchmark for multi-hop kinship reasoning. Each instance consists of a narrative describing family relations among named characters, a query asking for relation between two target characters, and a gold relation label. Solving an instance requires composing a sequence of intermediate kinship relations, for example composing \texttt{sister} and \texttt{father} to derive \texttt{aunt}. We randomly sample 3{,}073 instances from CLUTRR, stratified across reasoning chains of 2 to 10 hops.
We reserve 200 held-out 2-hop instances for ILP rule learning and validation. 
Longer chains are used only at evaluation time, where the goal is not to learn additional rules but to test whether the learned local rules can support iterative verification and repair under increasing reasoning depth.

\begin{table*}[ht]
\centering
\footnotesize
\renewcommand{\arraystretch}{1.2}
\begin{tabular}{@{} l l cccc @{}}
\toprule
& & \multicolumn{4}{c}{\textbf{Reasoning Chain Length (\# of hops)}} \\
\cmidrule(lr){3-6}
\textbf{Model} & \textbf{Reasoning Setups} & Short (2-3) & Medium (4-5) & Long (6-7) & Longer (8-10) \\
\midrule
\multirow{3}{*}{Qwen-3.5:4B}
  & Endogenous          & 56.95$\pm$1.73 & 43.74$\pm$3.49 & 25.91$\pm$3.45 & 19.09$\pm$2.00 \\
  & Exogenous            & \textbf{75.07}$\pm$1.33 & 70.94$\pm$2.90 & 56.67$\pm$4.06 & 46.72$\pm$4.35 \\
  & \textbf{\textsc{Popper}-ly (Ours)}   & 73.23$\pm$1.05 & \textbf{72.78}$\pm$3.87 & \textbf{67.36}$\pm$4.16 & \textbf{61.93}$\pm$5.33 \\
\addlinespace
\multirow{3}{*}{Gemma-4:E4B}
  & Endogenous          & 65.92$\pm$1.91 & 54.05$\pm$5.75 & 38.14$\pm$6.13 & 14.39$\pm$3.15 \\
  & Exogenous            & \textbf{78.35}$\pm$1.57 & 71.67$\pm$4.33 & 54.56$\pm$4.44 & 42.22$\pm$3.39 \\
  & \textbf{\textsc{Popper}-ly (Ours)}   & 77.84$\pm$1.95 & \textbf{74.68}$\pm$5.02 & \textbf{66.22}$\pm$3.93 & \textbf{62.74}$\pm$3.58 \\
\addlinespace
\multirow{3}{*}{LLaMA-3.2:3B}
  & Endogenous          & 39.48$\pm$2.79 & 29.68$\pm$1.90 & 28.67$\pm$7.17 & 18.72$\pm$5.26 \\
  & Exogenous            & 63.93$\pm$1.99 & 54.41$\pm$2.19 & 39.91$\pm$5.14 & 33.48$\pm$6.12 \\
  & \textbf{\textsc{Popper}-ly (Ours)}   & \textbf{70.36}$\pm$1.91 & \textbf{63.67}$\pm$2.09 & \textbf{52.32}$\pm$6.16 & \textbf{42.48}$\pm$7.62 \\
\midrule
\multirow{3}{*}{Claude-4.6 Sonnet}
  & Endogenous          & 86.75$\pm$0.88 & 74.15$\pm$6.35 & 64.93$\pm$9.45 & 56.92$\pm$2.83 \\
  & Exogenous            & 88.87$\pm$1.28 & 75.91$\pm$3.09 & 71.31$\pm$3.01 & 66.79$\pm$6.70 \\
  & \textbf{\textsc{Popper}-ly (Ours)}   & \textbf{89.63}$\pm$1.59 & \textbf{79.69}$\pm$3.70 & \textbf{73.33}$\pm$2.57 & \textbf{71.38}$\pm$7.71 \\
\addlinespace
\multirow{3}{*}{GPT-5.4}
  & Endogenous          & 89.23$\pm$0.94 & 78.06$\pm$2.69 & 68.15$\pm$4.72 & 59.91$\pm$5.56 \\
  & Exogenous            & \textbf{92.90}$\pm$1.39 & 80.44$\pm$3.89 & 71.75$\pm$6.45 & 69.53$\pm$3.45 \\
  & \textbf{\textsc{Popper}-ly (Ours)}   & 92.59$\pm$1.95 & \textbf{85.31}$\pm$4.41 & \textbf{79.62}$\pm$7.95 & \textbf{75.09}$\pm$5.35 \\
\bottomrule
\end{tabular}
\caption{Terminal accuracy (\%) on CLUTRR under multi-fold evaluation. \textit{Endogenous}: LLM reasoning via single-shot CoT. \textit{Exogenous}: LLM extracts facts and symbolic composition is performed using the learned rule table. \textsc{Popper}-ly: LLM generates CoT, and ILP-based verification and repair is applied to violated inferred steps. Best result in each row block is shown in \textbf{bold}.}
\label{tab:main}
\end{table*}

\paragraph{Research Question.}
\textit{Does step-level ILP-based verification and repair improve terminal reasoning accuracy over both endogenous CoT reasoning and fully exogenous symbolic composition, especially as compositional depth increases?}

\paragraph{Models.}
We evaluate following LLMs:
(1) \textit{Small local models.} Qwen-3.5:4B \cite{qwen35blog}, Gemma-4:E4B \cite{gemma4}, and LLaMA-3.2:3B \cite{touvron2023llama,llama3_2}. These models are run locally on an NVIDIA A6000 Ada GPU.
(2) \textit{Frontier API models.} Claude Sonnet 4.6 \cite{claude4_6} and GPT-5.4 \cite{gpt5_4}, accessed through their respective APIs.
All models receive the same structured CoT prompt with explicit formatting instructions. The final answer is constrained to one of the 21 CLUTRR relation labels, such as \texttt{father}, \texttt{aunt}, and \texttt{grandson-in-law}, in order to reduce answer-format errors that are unrelated to reasoning quality.

\paragraph{Metric.}
Our primary metric is \textit{terminal accuracy}, defined as the fraction of instances for which the model's extracted final answer exactly matches the gold relation label. We report terminal accuracy by model, reasoning configuration, and hop range. Our main quantity of interest is the accuracy gain of step-level symbolic patching over standard single-shot CoT.

\paragraph{Baselines.}
We compare Reason \textsc{Popper}-ly against two baselines.
(1) \textit{Endogenous reasoning (single-shot CoT).}
The LLM generates a full chain-of-thought trace and a final answer without any external verification or intervention. This is the standard CoT setup \cite{wei2022chain} and isolates the model's native compositional reasoning ability.
(2) \textit{Exogenous reasoning (symbolic composition pipeline).}
The LLM is used only to extract stated facts from the narrative \cite{schick2023toolformer,yao2022react}. We then build a relational graph over entities, identify a path between the query targets using breadth-first search, and compose relations along that path using the learned rule table $\mathcal{R}$. This baseline removes model-endogenous reasoning after extraction and therefore isolates the benefit of exact symbolic composition conditional on successful grounding. Its failures primarily reflect brittleness in fact extraction and graph construction rather than errors in symbolic composition itself.

\paragraph{Terminal Performance Analysis.}
Table~\ref{tab:main} reports terminal accuracy across five models, three reasoning configurations, and four hop bins. We highlight three findings.
\noindent \textit{(1) Step-level symbolic repair consistently improves over endogenous CoT.}
Across all models and all reasoning depths, Reason \textsc{Popper}-ly outperforms standard single-shot CoT. The improvements are especially large for smaller models on long chains. At 8-10 hops, Qwen improves from 19.09\% to 61.93\%, Gemma from 14.39\% to 62.74\%, and LLaMA from 18.72\% to 42.48\%. Even frontier models benefit materially: Claude Sonnet improves from 56.92\% to 71.38\%, and GPT-5.4 improves from 59.91\% to 75.09\%. These results indicate that a substantial portion of long-chain degradation arises from verifiable local composition errors that can be corrected after generation.
\noindent \textit{(2) The relative advantage of exogenous symbolic composition decreases as reasoning depth grows.}
A consistent crossover pattern appears across models. At short chains, the exogenous pipeline often matches or slightly exceeds Reason \textsc{Popper}-ly. This is expected: when the required reasoning depth is small, fact extraction is relatively easy and exact symbolic composition is highly effective. As chain length increases, however, Reason \textsc{Popper}-ly overtakes the exogenous baseline and the gap widens. At 8-10 hops, the advantage of Reason \textsc{Popper}-ly over exogenous reasoning ranges from 4.59 points for Claude Sonnet to 20.52 points for Gemma. This pattern suggests a shift in the dominant failure mode. For short chains, local reasoning mistakes dominate. For longer chains, extraction and graph-construction errors accumulate, which increasingly harms fully exogenous symbolic pipelines. In contrast, our method preserves the LLM's own extraction and intermediate structure whenever they are already usable, and intervenes only at reasoning steps that violate learned composition rules.
\noindent \textit{(3) Frontier models remain patchable.}
Claude Sonnet 4.6 and GPT-5.4 start from much stronger endogenous CoT baselines than the smaller local models, but both still exhibit substantial degradation as reasoning depth increases. The fact that step-level symbolic repair yields gains even at this scale shows that compositional reasoning failures are not restricted to weaker models. Instead, they remain present, detectable, and actionable even in strong frontier systems. This supports the broader view that lightweight symbolic guardrails can remain useful even when the base model is already highly capable.

\begin{figure}[h]
    \centering
    \includegraphics[width=0.65\linewidth]{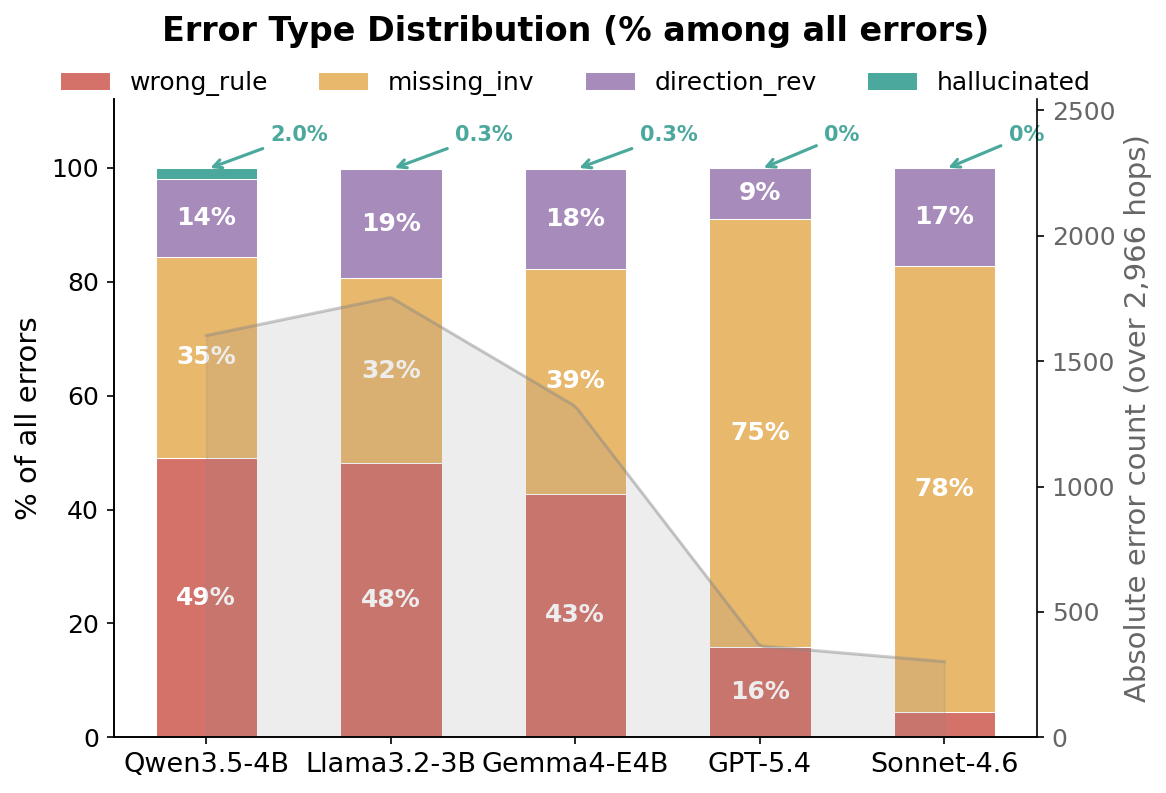}
    \caption{Distribution of diagnosed errors among reasoning hops.}
    \label{fig:error-type}
\end{figure}

\paragraph{Error Attribution Analysis.}
Figure~\ref{fig:error-type} shows the distribution of diagnosed error types among all detected reasoning errors. Two patterns are especially notable. 
(1) the error taxonomy reveals clear model-specific failure profiles rather than a single undifferentiated notion of reasoning error. Smaller local models exhibit a more balanced mix of \texttt{wrong\_rule}, \texttt{missing\_inv}, and \texttt{direction\_rev}, with \texttt{wrong\_rule} being the dominant category (43-49\%). In contrast, frontier models shift sharply toward \texttt{missing\_inv}, which accounts for 75\% of GPT-5.4 errors and 78\% of Sonnet-4.6 errors, while \texttt{wrong\_rule} drops to 16\% and 4\%, respectively. This suggests that \textit{scaling reduces core composition mistakes but does not eliminate structured reasoning failures; instead, the dominant failure mode shifts toward relation-direction normalization errors}. 
(2) \texttt{hallucinated} errors are nearly absent across all models (0-2\%), indicating that in this benchmark the main bottleneck is not unsupported premise invention, but incorrect symbolic composition and direction handling. These findings support the value of step-level symbolic diagnosis: \textit{it not only enables targeted repair, but also exposes qualitatively different reasoning weaknesses across model scales.}
\section{Related Work}
 
\textbf{Chain-of-thought reasoning} and failure modes.
Chain-of-thought prompting improves performance on many multi-step reasoning tasks by eliciting intermediate natural-language steps before final answer \cite{wei2022chain}. At the same time, a growing body of work shows that these intermediate traces should not be taken at face value. Prior studies find that CoT explanations can be unfaithful to the model's underlying decision process \cite{turpin2023language,lanhammeasuring}, and that compositional reasoning quality deteriorates rapidly as proof depth increases \cite{saparov2023language}. More broadly, process-level supervision has been motivated by the observation that final-answer correctness alone provides only a weak signal for reasoning quality \cite{lightman2023let}. Our work builds on this line of evidence, but focuses on different intervention: rather than supervising reasoning during training, we verify and repair explicit reasoning steps at inference time using learned rules.
\textbf{Neuro-symbolic reasoning with LLMs.}
A growing literature combines LLMs with symbolic reasoning systems to improve faithfulness, controllability, and logical correctness \cite{yang2023leandojo,hsiang2025leandojov,ijcai2025p1155,ijcai2025p1065,an2026logiex}. Logic-LM \cite{pan2023logic} translates natural-language problems into formal logic and delegates inference to an external solver. LINC \cite{olausson2023linc} similarly combines LLMs with first-order logic provers for logical reasoning. SymbCoT \cite{xu2024faithful} injects symbolic expressions directly into chain-of-thought traces and adds an LLM-based verification stage, while faithful reasoning approaches have also explored stronger structural constraints over intermediate reasoning \cite{lyu2023faithful}. These methods either rely on symbolic reasoning over the full problem or restructure the reasoning process globally. Our method operates on the model's own generated trace and intervenes only at individual steps that violate induced composition rules, preserving the remaining intermediate reasoning whenever possible.
\textbf{Self-correction and iterative refinement.}
Another related line of work studies whether LLMs can improve their own outputs through critique and revision. Self-Refine \cite{madaan2023self} iteratively feeds model-generated feedback back into the model for refinement, and CRITIC \cite{gou2024critic} augments revision with tool-interactive critique. These methods are general and often effective in practice, but they typically revise outputs through broad regeneration rather than localized, rule-grounded correction. Moreover, recent evidence suggests that LLMs cannot reliably self-correct reasoning in the absence of strong external feedback \cite{huang2024large,chen2025sim911}. Our method uses exactly such external feedback, but in a symbolic form: the model is not asked to discover its own mistake, only to continue from a corrected prefix that has been externally verified and repaired.
\textbf{Inductive logic programming.}
ILP learns logic programs from examples \cite{muggleton1991inductive}. Modern systems such as Popper \cite{cropper2021learning} perform hypothesis-driven search with efficient pruning from failed candidates, making them practical for learning compact symbolic rule sets. Recent work has also explored interactions between LLMs and ILP, for example by using LLMs to generate supervision for logical rule induction \cite{de2025inductive}. Our use of ILP is different in both role and timing. We do not use ILP as a standalone solver, nor as a training-time signal for updating the language model. Instead, we use ILP to induce a symbolic composition table offline and deploy the learned rules at inference time as a step-level verifier and repair mechanism over reasoning traces.
\section{Discussion, Limitation, and Future Work}
 
\textbf{Domain specificity.}
Our evaluation focuses on kinship reasoning, where relation composition is well defined and grounding from natural language to structured predicates is relatively tractable through templated parsing. This makes CLUTRR a useful testbed for studying whether symbolic verification can improve step-level reasoning, but it also limits the immediate generality of our findings. In domains with more ambiguous predicate structures, such as legal reasoning or scientific argumentation, both the grounding process and the induced rule space would be substantially more complex. 
\textbf{Assumptions about explicit reasoning.}
Our method assumes that the explicit CoT trace remains causally relevant to the model's final answer, so that editing an intermediate step can redirect the downstream conclusion. This assumption is not guaranteed. Prior work has shown that CoT explanations can function as post-hoc rationalizations rather than faithful accounts of the model's underlying computation \cite{turpin2023language,lanhammeasuring}. If the model has effectively committed to an answer before the full reasoning trace is produced, then patching intermediate steps may have limited influence on the final prediction. Our empirical results suggest that the assumption is at least partially valid in the evaluated setting, since symbolic patching yields consistent gains in terminal accuracy. 


\section{Conclusion}
 
We presented Reason \textsc{Popper}-ly, a framework for improving LLM CoT reasoning through ILP-based step-level verification and repair. The method learns symbolic relation composition rules offline, then applies them at inference time to detect violated reasoning steps, diagnose their error type, patch invalid compositions, and regenerate only the affected suffix of the trace. In this way, the framework preserves the model's own reasoning wherever it is already valid while introducing symbolic correction only where it is needed.
Experiments across multiple LLMs and reasoning depths show that Reason \textsc{Popper}-ly consistently improves terminal accuracy over standard CoT prompting. 
The gains become more pronounced as reasoning chains grow longer, where local composition errors accumulate and increasingly degrade unassisted model reasoning. 
Overall, these results suggest that symbolic reasoning support need not fully replace model-generated reasoning to be useful. In compositional settings, a lightweight verifier that selectively repairs invalid steps can offer a practical middle ground between unconstrained CoT and full symbolic delegation.

\newpage
\bibliographystyle{plainnat}
\bibliography{refs}

@article{wei2022chain,
  title={Chain-of-thought prompting elicits reasoning in large language models},
  author={Wei, Jason and Wang, Xuezhi and Schuurmans, Dale and Bosma, Maarten and Xia, Fei and Chi, Ed and Le, Quoc V and Zhou, Denny and others},
  journal={Advances in neural information processing systems},
  volume={35},
  pages={24824--24837},
  year={2022}
}

@article{muggleton1991inductive,
  title={Inductive logic programming},
  author={Muggleton, Stephen},
  journal={New generation computing},
  volume={8},
  number={4},
  pages={295--318},
  year={1991},
  publisher={Springer}
}

@article{cropper2021learning,
  title={Learning programs by learning from failures},
  author={Cropper, Andrew and Morel, Rolf},
  journal={Machine Learning},
  volume={110},
  number={4},
  pages={801--856},
  year={2021},
  publisher={Springer}
}

@article{turpin2023language,
  title={Language models don't always say what they think: Unfaithful explanations in chain-of-thought prompting},
  author={Turpin, Miles and Michael, Julian and Perez, Ethan and Bowman, Samuel},
  journal={Advances in Neural Information Processing Systems},
  volume={36},
  pages={74952--74965},
  year={2023}
}

@inproceedings{
saparov2023language,
title={Language Models Are Greedy Reasoners: A Systematic Formal Analysis of Chain-of-Thought},
author={Abulhair Saparov and He He},
booktitle={The Eleventh International Conference on Learning Representations },
year={2023},
url={https://openreview.net/forum?id=qFVVBzXxR2V}
}

@inproceedings{sinha2019clutrr,
  title={CLUTRR: A diagnostic benchmark for inductive reasoning from text},
  author={Sinha, Koustuv and Sodhani, Shagun and Dong, Jin and Pineau, Joelle and Hamilton, William L},
  booktitle={Proceedings of the 2019 Conference on Empirical Methods in Natural Language Processing and the 9th International Joint Conference on Natural Language Processing (EMNLP-IJCNLP)},
  pages={4506--4515},
  year={2019}
}

@inproceedings{xu2024faithful,
  title={Faithful logical reasoning via symbolic chain-of-thought},
  author={Xu, Jundong and Fei, Hao and Pan, Liangming and Liu, Qian and Lee, Mong-Li and Hsu, Wynne},
  booktitle={Proceedings of the 62nd Annual Meeting of the Association for Computational Linguistics (Volume 1: Long Papers)},
  pages={13326--13365},
  year={2024}
}

@inproceedings{pan2023logic,
  title={Logic-lm: Empowering large language models with symbolic solvers for faithful logical reasoning},
  author={Pan, Liangming and Albalak, Alon and Wang, Xinyi and Wang, William},
  booktitle={Findings of the Association for Computational Linguistics: EMNLP 2023},
  pages={3806--3824},
  year={2023}
}

@article{lanhammeasuring,
  title={Measuring faithfulness in chain-of-thought reasoning},
  author={Lanham, Tamera and Chen, Anna and Radhakrishnan, Ansh and Steiner, Benoit and Denison, Carson and Hernandez, Danny and Li, Dustin and Durmus, Esin and Hubinger, Evan and Kernion, Jackson and others},
  journal={arXiv preprint arXiv:2307.13702},
  year={2023}
}

@inproceedings{de2025inductive,
  title={Inductive learning of logical theories with llms: A expressivity-graded analysis},
  author={de Souza, Jo{\~a}o Pedro Gandarela and Carvalho, Danilo and Freitas, Andr{\'e}},
  booktitle={Proceedings of the AAAI Conference on Artificial Intelligence},
  volume={39},
  pages={23752--23759},
  year={2025}
}

@article{madaan2023self,
  title={Self-refine: Iterative refinement with self-feedback},
  author={Madaan, Aman and Tandon, Niket and Gupta, Prakhar and Hallinan, Skyler and Gao, Luyu and Wiegreffe, Sarah and Alon, Uri and Dziri, Nouha and Prabhumoye, Shrimai and Yang, Yiming and others},
  journal={Advances in neural information processing systems},
  volume={36},
  pages={46534--46594},
  year={2023}
}

@inproceedings{
gou2024critic,
title={{CRITIC}: Large Language Models Can Self-Correct with Tool-Interactive Critiquing},
author={Zhibin Gou and Zhihong Shao and Yeyun Gong and yelong shen and Yujiu Yang and Nan Duan and Weizhu Chen},
booktitle={The Twelfth International Conference on Learning Representations},
year={2024},
url={https://openreview.net/forum?id=Sx038qxjek}
}

@inproceedings{
huang2024large,
title={Large Language Models Cannot Self-Correct Reasoning Yet},
author={Jie Huang and Xinyun Chen and Swaroop Mishra and Huaixiu Steven Zheng and Adams Wei Yu and Xinying Song and Denny Zhou},
booktitle={The Twelfth International Conference on Learning Representations},
year={2024},
url={https://openreview.net/forum?id=IkmD3fKBPQ}
}

@article{schick2023toolformer,
  title={Toolformer: Language models can teach themselves to use tools},
  author={Schick, Timo and Dwivedi-Yu, Jane and Dess{\`\i}, Roberto and Raileanu, Roberta and Lomeli, Maria and Hambro, Eric and Zettlemoyer, Luke and Cancedda, Nicola and Scialom, Thomas},
  journal={Advances in neural information processing systems},
  volume={36},
  pages={68539--68551},
  year={2023}
}

@inproceedings{yao2022react,
  title={React: Synergizing reasoning and acting in language models},
  author={Yao, Shunyu and Zhao, Jeffrey and Yu, Dian and Du, Nan and Shafran, Izhak and Narasimhan, Karthik R and Cao, Yuan},
  year={2022},
  booktitle={The eleventh international conference on learning representations}
}

@misc{qwen35blog,
    title = {Qwen3.5: Accelerating Productivity with Native Multimodal Agents},
    url = {https://qwen.ai/blog?id=qwen3.5},
    author = {{Qwen Team}},
    month = {February},
    year = {2026}
}

@misc{gemma4,
  title        = {Gemma 4: Lightweight Open Models for Reasoning and Agentic Systems},
  author       = {{Google DeepMind}},
  year         = {2026},
  howpublished = {\url{https://deepmind.google/models/gemma/gemma-4/}},
  note         = {Accessed: 2026-04-07}
}

@misc{llama3_2,
  title        = {Llama 3.2: Open Foundation and Multimodal Models},
  author       = {{Meta AI}},
  year         = {2024},
  howpublished = {\url{https://ai.meta.com/llama/}},
  note         = {Accessed: 2026-04-07}
}

@article{touvron2023llama,
  title={Llama: Open and efficient foundation language models},
  author={Touvron, Hugo and Lavril, Thibaut and Izacard, Gautier and Martinet, Xavier and Lachaux, Marie-Anne and Lacroix, Timoth{\'e}e and Rozi{\`e}re, Baptiste and Goyal, Naman and Hambro, Eric and Azhar, Faisal and others},
  journal={arXiv preprint arXiv:2302.13971},
  year={2023}
}

@misc{claude4_6,
  title        = {Claude 4.6},
  author       = {{Anthropic}},
  year         = {2026},
  howpublished = {\url{https://www.anthropic.com/news/claude-sonnet-4-6}},
  note         = {Accessed: 2026-04-07}
}

@misc{gpt5_4,
  title        = {GPT-5.4},
  author       = {{OpenAI}},
  year         = {2026},
  howpublished = {\url{https://openai.com/index/introducing-gpt-5-4/}},
  note         = {Accessed: 2026-04-07}
}

@inproceedings{olausson2023linc,
  title={LINC: A neurosymbolic approach for logical reasoning by combining language models with first-order logic provers},
  author={Olausson, Theo and Gu, Alex and Lipkin, Ben and Zhang, Cedegao and Solar-Lezama, Armando and Tenenbaum, Joshua and Levy, Roger},
  booktitle={Proceedings of the 2023 Conference on Empirical Methods in Natural Language Processing},
  pages={5153--5176},
  year={2023}
}

@inproceedings{lightman2023let,
  title={Let's verify step by step},
  author={Lightman, Hunter and Kosaraju, Vineet and Burda, Yuri and Edwards, Harrison and Baker, Bowen and Lee, Teddy and Leike, Jan and Schulman, John and Sutskever, Ilya and Cobbe, Karl},
  booktitle={The twelfth international conference on learning representations},
  year={2023}
}

@inproceedings{lyu2023faithful,
  title={Faithful chain-of-thought reasoning},
  author={Lyu, Qing and Havaldar, Shreya and Stein, Adam and Zhang, Li and Rao, Delip and Wong, Eric and Apidianaki, Marianna and Callison-Burch, Chris},
  booktitle={Proceedings of the 13th International Joint Conference on Natural Language Processing and the 3rd Conference of the Asia-Pacific Chapter of the Association for Computational Linguistics (Volume 1: Long Papers)},
  pages={305--329},
  year={2023}
}

@inproceedings{ijcai2025p1155,
  title     = {Empowering LLMs with Logical Reasoning: A Comprehensive Survey},
  author    = {Cheng, Fengxiang and Li, Haoxuan and Liu, Fenrong and van Rooij, Robert and Zhang, Kun and Lin, Zhouchen},
  booktitle = {Proceedings of the Thirty-Fourth International Joint Conference on
               Artificial Intelligence, {IJCAI-25}},
  publisher = {International Joint Conferences on Artificial Intelligence Organization},
  editor    = {James Kwok},
  pages     = {10400--10408},
  year      = {2025},
  month     = {8},
  note      = {Survey Track},
  doi       = {10.24963/ijcai.2025/1155},
  url       = {https://doi.org/10.24963/ijcai.2025/1155},
}

@article{yang2023leandojo,
  title={Leandojo: Theorem proving with retrieval-augmented language models},
  author={Yang, Kaiyu and Swope, Aidan and Gu, Alex and Chalamala, Rahul and Song, Peiyang and Yu, Shixing and Godil, Saad and Prenger, Ryan J and Anandkumar, Animashree},
  journal={Advances in Neural Information Processing Systems},
  volume={36},
  pages={21573--21612},
  year={2023}
}

@inproceedings{
hsiang2025leandojov,
title={LeanDojo-v2: A Comprehensive Library for {AI}-Assisted Theorem Proving in Lean},
author={Ryan Hsiang and William Adkisson and Robert Joseph George and Anima Anandkumar},
booktitle={The 5th Workshop on Mathematical Reasoning and AI at NeurIPS 2025},
year={2025},
url={https://openreview.net/forum?id=tnx1VvrcAn}
}

@inproceedings{chen2022cityspec,
  title={Cityspec: An intelligent assistant system for requirement specification in smart cities},
  author={Chen, Zirong and Li, Isaac and Zhang, Haoxiang and Preum, Sarah and Stankovic, John A and Ma, Meiyi},
  booktitle={2022 IEEE International Conference on Smart Computing (SMARTCOMP)},
  pages={32--39},
  year={2022},
  organization={IEEE}
}

@article{chen2023cityspec,
  title={CitySpec with shield: A secure intelligent assistant for requirement formalization},
  author={Chen, Zirong and Li, Isaac and Zhang, Haoxiang and Preum, Sarah and Stankovic, John A and Ma, Meiyi},
  journal={Pervasive and Mobile Computing},
  volume={92},
  pages={101802},
  year={2023},
  publisher={Elsevier}
}

@inproceedings{ijcai2025p1065,
  title     = {LogiDebrief: A Signal-Temporal Logic Based Automated Debriefing Approach with Large Language Models Integration},
  author    = {Chen, Zirong and An, Ziyan and Reynolds, Jennifer and Mullen, Kristin and Maritini, Stephen and Ma, Meiyi},
  booktitle = {Proceedings of the Thirty-Fourth International Joint Conference on
               Artificial Intelligence, {IJCAI-25}},
  publisher = {International Joint Conferences on Artificial Intelligence Organization},
  editor    = {James Kwok},
  pages     = {9582--9590},
  year      = {2025},
  month     = {8},
  note      = {AI and Social Good},
  doi       = {10.24963/ijcai.2025/1065},
  url       = {https://doi.org/10.24963/ijcai.2025/1065},
}

@inproceedings{an2026logiex,
  title={LogiEx: Integrating Formal Logic and LLMs for Explainable Transit Planning},
  author={An, Ziyan and Wang, Xia and Baier, Hendrik and Chen, Zirong and Dubey, Abhishek and Johnson, Taylor T and Sprinkle, Jonathan and Mukhopadhyay, Ayan and Ma, Meiyi},
  booktitle={2026 ACM/IEEE 17th International Conference on Cyber-Physical Systems (ICCPS)},
  pages={88--99},
  year={2026},
  organization={IEEE}
}

@inproceedings{chen2025sim911,
  title={Sim911: towards effective and equitable 9-1-1 dispatcher training with an llm-enabled simulation},
  author={Chen, Zirong and Chason, Elizabeth and Mladenovski, Noah and Wilson, Erin and Mullen, Kristin and Martini, Stephen and Ma, Meiyi},
  booktitle={Proceedings of the AAAI Conference on Artificial Intelligence},
  volume={39},
  pages={27896--27904},
  year={2025}
}

\end{document}